\documentclass{article}

\PassOptionsToPackage{numbers, compress}{natbib}
\usepackage[preprint]{neurips_2020}

\usepackage[utf8]{inputenc} % allow utf-8 input
\usepackage[T1]{fontenc}    % use 8-bit T1 fonts
\usepackage{hyperref}       % hyperlinks
\usepackage{url}            % simple URL typesetting
\usepackage{booktabs}       % professional-quality tables
\usepackage{amsfonts}       % blackboard math symbols
\usepackage{nicefrac}       % compact symbols for 1/2, etc.
\usepackage{microtype}      % microtypography
\usepackage{graphicx}
\usepackage{amsmath}
\usepackage{amsthm}
\usepackage{amssymb}
\usepackage{algorithm}
\usepackage{algorithmicx}
\usepackage[noend]{algpseudocode}
\usepackage{subfigure}
\usepackage{array}

\newcommand{\argmax}{\operatornamewithlimits{arg\,max}}
\newcommand{\argmin}{\operatornamewithlimits{arg\,min}}

\newtheorem{remark}{Remark}

\def\xx{\boldsymbol x}

\def\ss{\boldsymbol s}

\title{Neural Architecture Optimization with Graph VAE}

\author{
Jian Li$^{1,2}$, Yong Liu$^{1,2}$\thanks{Corresponding author}~~, Jiankun Liu$^{1,2}$, Weiping Wang$^{1,2}$\\
$ ^1$Institute of Information Engineering, Chinese Academy of Sciences\\
$ ^2$School of Cyber Security, University of Chinese Academy of Sciences\\
\texttt{\{lijian9026,liuyong,liujiankun,wangweiping\}@iie.ac.cn}
}

\begin{document}

\maketitle

\begin{abstract}
  Due to their high computational efficiency on a continuous space, gradient optimization methods have shown great potential in the neural architecture search (NAS) domain. The mapping of network representation from the discrete space to a latent space is the key to discovering novel architectures, however, existing gradient-based methods fail to fully characterize the networks. In this paper, we propose an efficient NAS approach to optimize network architectures in a continuous space, where the latent space is built upon variational autoencoder (VAE) and graph neural networks (GNN). The framework jointly learns four components: the encoder, the performance predictor, the complexity predictor and the decoder in an end-to-end manner. The encoder and the decoder belong to a graph VAE, mapping architectures between continuous representations and network architectures. The predictors are two regression models, fitting the performance and computational complexity, respectively. Those predictors ensure the discovered architectures characterize both excellent performance and high computational efficiency. Extensive experiments demonstrate our framework not only generates appropriate continuous representations but also discovers powerful neural architectures.
\end{abstract}

\section{Introduction}
Deep neural networks have achieved great success in various domains due to powerful hierarchical feature representations generated by well-designed neural architectures.
For example, convolution neural networks (CNN) have evolved from AlexNet \cite{krizhevsky2012imagenet} into several architectures, including VGG \cite{simonyan2014very}, Inception \cite{szegedy2015going}, ResNet \cite{he2016deep} and DenseNet \cite{huang2017densely}.
However, the design of network architectures highly depends on the artificial experience and extremely inefficient, which promotes the research on \textit{Neural Network Search (NAS)} \cite{zoph2016neural,baker2016designing,real2017large,zoph2018learning,luo2018neural,zhou2019bayesnas}.
NAS discovers promising architectures via optimization on the search space, outperforming handcrafted networks \cite{zoph2018learning,luo2018neural,real2019regularized}.  
There are two key issues: 

1) \textbf{Representation}.
Typically, neural architectures are represented in a high dimensional discrete space according to the candidate operators and hyperparameters, such as graph representations in reinforcement learning (RL) \cite{zoph2018learning} and evolutionary algorithms (EA) \cite{real2017large,real2019regularized}.
However, discrete representations of architecture lead to enormous search space and hard to optimize.
Researchers tried to encode network architectures to a continuous space, where the mapping is the key to finding novel architectures in the continuous space.
Bayesian optimization (BO) \cite{kandasamy2018neural} and gradient-based methods \cite{luo2018neural,liu2018darts} encode architectures into a continuous space and then optimize architectures in the space.

2) \textbf{Optimization}. Based on the discrete space of different layers in architectures, RL methods \cite{zoph2016neural,zoph2018learning} use policy optimization to choose a layer's type and corresponding hyperparameters.
EA approaches \cite{real2017large,gaier2019weight,real2019regularized} utilize genetic algorithms to optimize both the neural architectures and their weights.
Typical Bayesian optimization (BO) approaches \cite{kandasamy2018neural,zhou2019bayesnas} are based on Gaussian process and more suitable for low-dimensional continuous optimization problems, thus they first map the discrete representations to continuous representations.
Gradient-based methods optimize \cite{luo2018neural,liu2018darts,cai2018proxylessnas,xie2018snas} the network architecture on the continuous space w.r.t performance of architectures.

\begin{figure}[t]
  \centering
  \includegraphics[width=\textwidth]{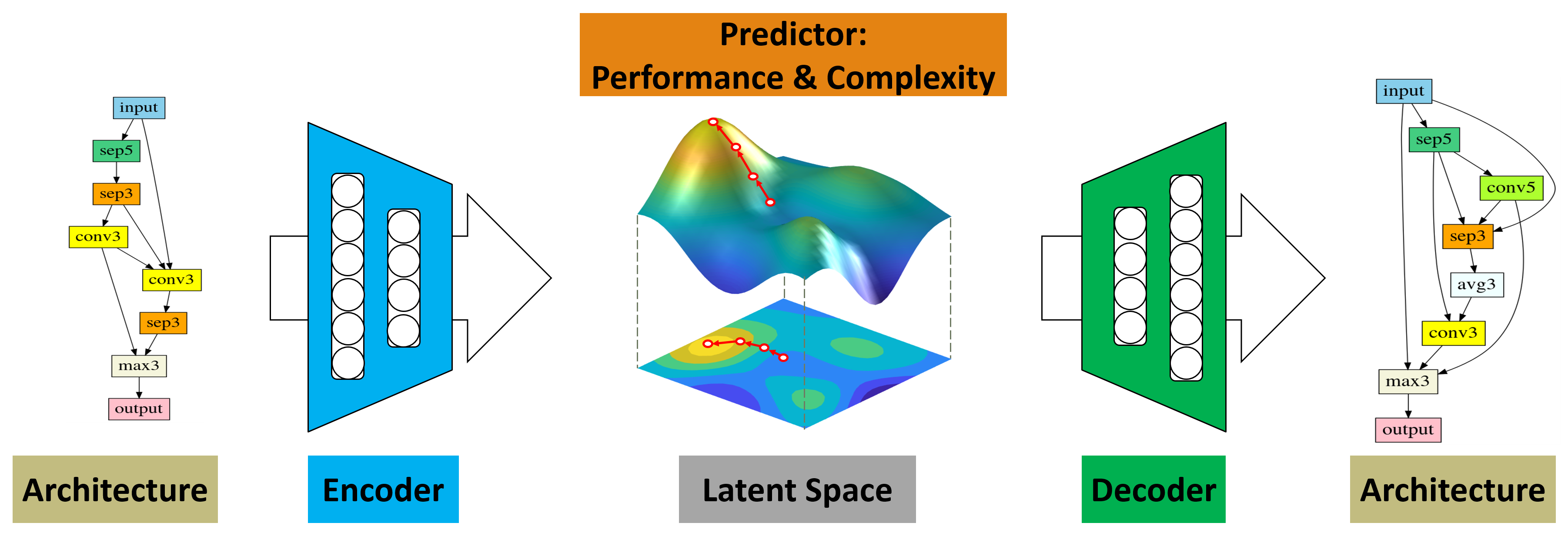}
  \caption{The learning framework of \texttt{NGAE}.
  The neural architecture is mapped into a continuous representation by the encoder network.
  Using stochastic gradient ascent, the continuous representation is optimized via maximizing the predictor, combining performance and complexity predictor.
  The optimized continuous representation is decoded out to a network architecture using the decoder.}
  \label{fig.architecture}
\end{figure}

Gradient-based approaches have drawn much attention because of their continuous optimization and high efficiency.
However, existing gradient methods have two weaknesses: 1) Graph representations use continuous relaxation \cite{liu2018darts} or sequential embeddings \cite{luo2018neural} focus on the structure but fail to extract computations in the neural architectures. 2) Existing gradient methods pursue architectures with better performances but ignore the computational efficiency.

In this paper, we propose a gradient-based Neural architecture optimization with Graph variational AutoEncoder (\texttt{NGAE}), which aims to find architectures with both preferable performance and high computational efficiency.
As shown in Figure \ref{fig.architecture}, three are three parts in the framework \texttt{NGAE}: 
1) The \textit{encoder} is to map the architectures to a latent space, where we use graph neural networks (GNN) to obtain continuous representations of architectures. 
2) The \textit{predictor} consists of two regression models fitting the performance and the computational complexity in the latent space. 
3) The \textit{decoder} is to recover neural architectures from the continuous representations.

% In the training process, we jointly learn the encoder, the decoder and the predictors (both performance and complexity) in an end-to-end manner.
% Those components are constructed by multiple layers perceptrons (MLPs), thus \texttt{NGAE} optimizes the parameters of those MLPs via back propagation by minimizing the objective.
% The training objective covers the Kullback-Leibler (KL) divergence, the reconstruction loss, the squared loss of both performance predictor and complexity predictor.

% In the inference process, the inference objective aims to enlarge the output of the performance predictor and reduce the output of the complexity predictor, thus we merge the performance predictor and the complexity predictor as one into one predictor with negative linear relationship.
% \texttt{NGAE} optimizes continuous representations by maximizing the inference objective and then use the decoder to decode optimized connections representations into architectures.

\section{Related Work}
\subsection{Variational Graph Autoencoder}
Variational graph autoencoder is a kind of graph generative model using \textbf{variational autoencoder (VAE)} on graph embeddings.
VAE aims to learn both the decoder $p_\theta(\xx|\ss)$ and the encoder $q_\phi(\ss|\xx)$ via maximizing the variational evidence lower bound (ELBO) \cite{kingma2013auto}, which is stated as
\begin{align}
  \label{eq.lower_bound}
  \mathcal{L}({\boldsymbol \phi}, {\boldsymbol \theta}; \xx) = \mathbb{E}_{z \sim q_{\boldsymbol \phi}(\ss|\xx)} \left[\log p_{\boldsymbol \theta}(\xx | \ss)\right] - D_\text{KL} \left[q_{\boldsymbol \phi}(\ss | \xx) ~ || ~ p_{\boldsymbol \theta}(\ss)\right].
\end{align}
The first term estimates the reconstruction accuracy which reflects the similarity between the primal graphs and the decoded graphs.
The KL divergence term $D_\text{KL}$ regularizes the latent space.

NAO employed two LSTM models as the encoder and the decoder, which generates sequential representations of neural architectures \cite{luo2018neural}.
Graphrnn used a second-level RNN to capture edge dependencies \cite{you2018graphrnn}.
To better extract characteristics of graphs, recent work began to incorporate VAE and \textbf{graph neural networks (GNNs)}.
Using graph convolutional networks (GCN), \cite{li2018learning} represented neural architectures with the synchronous message passing scheme.
In contrast, D-VAE \cite{zhang2019d} employed asynchronous message passing schema as known from directed acyclic graphs (DAGs), to extract representations of nodes and edges in neural architectures.
Using GNNs, VS-GAE \cite{friede2019variational} was specialized in the representations of architectures with varying lengths.

\subsection{Gradient-based NAS methods}
While RL and EA methods search architectures in a discrete space, gradient-based methods use continuous representations to enable differentiable optimization.
NAO \cite{luo2018neural} encodes the neural architectures into continuous representations using VAE and discover new powerful architectures via stochastic gradient ascent by maximizing the predictor.
DARTS \cite{liu2018darts} uses a continuous relaxation to enable gradient-based optimization, where the graph is represented by the convex combination of the possibilities of all edges.
Both SNAS \cite{xie2018snas} and ProxylessNAS \cite{cai2018proxylessnas} optimize a parametrized
distribution over the possible operations.
NASP \cite{yao2020efficient} searches architectures in a differentiable space with a discrete constraint on architectures and a regularizer on model complexity.

\section{Learning Framework}
Let $(\xx, y, z) \in \mathcal{X} \times \mathcal{Y} \times \mathcal{Z}$ be a triplet of the architecture, its performance (e.g. classification accuracy of the network) and its computational complexity (e.g. the amount of computations), where $\mathcal{X}$ is the search space and $\mathcal{Y}, \mathcal{Z} \in \mathbb{R}^+$ are performance measure and computational measure, respectively.
The learning framework consists of four components: the encoder $E$, the performance predictor $f_{perf}$, the complexity predictor $f_{comp}$ and the decoder $D$.
The encoder $E: \mathcal{X} \to \mathcal{S}$ and the decoder $D: \mathcal{S} \to \mathcal{X}$ are used to convert representations between networks search space $\mathcal{X}$ and a continuous space $\mathcal{S}$.
Based on the latent space $\mathcal{S}$, the performance predictor $f_{perf}: \mathcal{S} \to \mathcal{Y}$ is used to fit the performance of network architectures, while the complexity predictor $f_{comp}: \mathcal{S} \to \mathcal{Z}$ is used to fit the computational complexity of network architectures,
We first train the encoder, two predictors and the decoder in an end-to-end manner, and then use them to infer new architectures. 

\subsection{Encoder}
Neural networks pass computations step-wise without circles, therefore networks are directed acyclic graphs (DAGs), where layers correspond to nodes in DAG and mappings between layers are edges in DAG.
To fully extract underlying characteristics of neural architectures, we employ a specialized GNN built for DAG \cite{zhang2019d}, which produces more powerful representations than traditional GNN.
DAG encoders focus on \textit{computations} rather than the \textit{structure}, dramatically reducing the search space.

We let the encoder be $E: \mathcal{X} \to \mathcal{S}$, where the encoder maps neural networks to continuous representations.
The continuous representation of a network architecture $\xx \in \mathcal{X}$ is built up in following steps: $\xx \to h_{out} \to \ss$, where the architecture $\xx$ is first encoded as a discrete vector $h_{out}$ by GNN and then mapped to a continuous representation $\ss$ by a variational autoencoder (VAE).

\subsubsection{Discrete Representations: $\xx \to h_{out}$}
We consider the asynchronous message passing schema in DAG where the computation of a node is carried out until it receives all incoming messages.
Therefore, the representations of nodes need to embed both the node type and incoming edges.
The discrete embedding of node $v$ is given by:
\begin{align}
  \label{eq.graph_embedding}
  h_v = \mathcal{U} \big(T_v, ~ \mathcal{A}(\{h_u: u \to v\}) \big),
\end{align}
where $T_v$ is the type of node $v$, represented as a one-hot vector.
$u \to v$ denotes the directed edge from node $u$ to node $v$, thus $\{h_u: u \to v\}$ is the set of $v$'s predecessors, which are determined by incoming edges.
The aggregate function $\mathcal{A}$ is used to aggregate all predecessors of $v$ and $\mathcal{U}$ fuses the representations of the node type and incoming edges.
There are no predecessors for the input node, so the aggregation for the input node is an all-zeros vector.

Following the topological ordering in the DAG, we iteratively compute the discrete vector for each node and finally the computations end in the output node $h_{out}$, where $h_{out}$ is used to represent the neural network $\xx$.
Although the graph embedding $h_{out}$ extracts structural information of neural networks, it is hard to optimized due to the discrete representations.

\begin{remark}
  Due to the asynchronous messages passing schema of DAG, the incoming edges $\{h_u: u \to v\}$ are invariant to nodes order permutations, so the aggregate function $\mathcal{A}$ should be invariant to the order of inputs.
  Meanwhile, to ensure the correspondence of $\xx$ and $h_{out}$, the aggregate function $\mathcal{A}$ and the update function $\mathcal{U}$ should be injective.
  In this paper, we let $\mathcal{A}$ be a gated sum and $\mathcal{U}$ be a gated recurrent unit (GRU).
\end{remark}

\subsubsection{Continuous Representations: $h_{out} \to \ss$}
We then introduce variational autoencoder (VAE) to obtain continuous representations of neural networks. For the sake of simplification, we let the variational approximate posterior $q_{\boldsymbol \phi}(\ss | h_{out})$ be a multivariate Gaussian with a diagonal structure:
\begin{align*}
  q_{\boldsymbol \phi}(\ss | h_{out}) = \mathcal{N}(\ss; {\boldsymbol \mu}, {\boldsymbol \sigma}^2 \mathbf{I})
\end{align*}
where we use two multilayer perceptrons (MLPs) to generate the mean ${\boldsymbol \mu}$ and the logarithm of variance $\log ({\boldsymbol \sigma^2})$, respectively.

To generate samples in the latent space $\mathcal{S}$ efficiently and accelerate the computation of the KL term in the variational lower bound \eqref{eq.lower_bound}, we let the prior be $p_\theta(\ss) = \mathcal{N}({\boldsymbol 0}, \mathbf{I})$. 
The maximization of the variational lower bound means to minimize the Kullback-Leibler divergence between the posterior $q_\phi(\ss | \xx)$ and the prior $p_\theta(\ss)$, then it holds \cite{kingma2013auto}:
\begin{equation}
  \label{eq.kl_divergence}
  \begin{aligned}
    - D_\text{KL} \big[q_\phi(\ss|h_{out}) ~||~ p_\theta(\ss)\big] 
    % &= \int q_\phi(\ss|h_{out}) \big[\log p_\theta(\ss) - \log q_\phi(\ss|h_{out})\big] ~ d \ss \\
    % &= \int \mathcal{N}(\ss; {\boldsymbol \mu}, {\boldsymbol \sigma}^2 \mathbf{I}) \big[ \log \mathcal{N}({\boldsymbol 0}, \mathbf{I}) - \log \mathcal{N}(\ss; {\boldsymbol \mu}, {\boldsymbol \sigma}^2 \mathbf{I}) \big] ~ d \ss \\
    &= \frac{1}{2} \sum_{j=1}^J \big[ 1 + \log ({\boldsymbol \sigma}_j^2) - {\boldsymbol \mu}_j^2 - {\boldsymbol \sigma}_j^2\big]
  \end{aligned}  
\end{equation}
where $J$ is the dimensionality of the latent space $\mathcal{S}$.
Therefore, we can directly sample points in the latent space from the standard normalization distribution and connect those two MLPs $f_{mean}$ and $f_{logvar}$ to the variational lower bound \eqref{eq.lower_bound}.

\subsection{Predictor}
The architecture with its performance and complexity $(\xx, y, z) \in \mathcal{X} \times \mathcal{Y} \times \mathcal{Z}$ is converted to $(\ss, y, z) \in \mathcal{S} \times \mathcal{Y} \times \mathcal{Z}$ after encoding.
The performance predictor $f_{perf}: \mathcal{S} \to \mathcal{Y}$ is a regression model to fit the performance of neural networks, where the performance rating is the accuracy for classification tasks and $\mathcal{Y} = [0, 1].$
The complexity predictor $f_{comp}: \mathcal{S} \to \mathcal{Z}$ is another regression model to fit the computational complexity of neural networks, where we regularize the computational complexity to $[0, 1].$ 
Those two predictors can be implemented by any nonlinear regression models, such as kernel ridge regression (KRR), Gaussian process regression and multilayer perceptron (MLP).
To facilitate the training of the predictor together with the encoder/decoder, we use two two-hidden-layers neural network (MLP) as the performance predictor $f_{perf}$ and the complexity predictor $f_{comp}$, respectively. 
For classification, we use Sigmoid as the activation function of the last layers, to ensure the predicted accuracy and complexity valid.

In the training process, we train two predictors w.r.t the squared loss $[f_{perf}(\ss) - y]^2, [f_{comp}(\ss) - z]^2,$ where $\ss$ is the latent representation of an architecture, $y$ is the corresponding performance and $z$ is the corresponding computational complexity.
The optimization objective of predictors is to minimize the empirical loss on the training data $(X, Y, Z)$, written as
\begin{align}
  \label{eq.squared_loss}
  \mathcal{L}_{pred} = \sum_{(\xx, y, z) \in (X, Y, Z)} \big[ f_{perf}(\ss) - y \big]^2 + \big[ f_{comp}(\ss) - z \big]^2.
\end{align} 

In the inference process, we directly optimize the neural architectures in the latent space based on the performance predictor $f_{perf}$ and the complexity predictor $f_{comp}$, where the objective is to maximize the predicted performance and minimize the predicted complexity.
Then, we use the decoder $D$ to decode the network architecture from the latent space $\mathcal{S}$ to the search space $\mathcal{X}$.

\subsection{Decoder}
We utilize the generative model of the graph autoencoder $D: \mathcal{S} \to \mathcal{X}$ to decode latent representations to DAGs (the network architectures), where the decoder also exerts the asynchronous message passing schema to construct valid DAGs. 
Using another MLP, we decode the continuous representation $\ss$ to $h_0$ as the initial discrete representation of the graph $\xx$.
We generate nodes in the graph until the output node $h_{out}$ is generated or reach the max allowed layers, where the representation of each node $h_v$ and the neural network architecture are generated as the following steps:

\textbf{1) Add a node.} Based on the representation of last generated node $h_v'$, we use an MLP followed by a softmax to determine the probability of the types of $h_v$, where we sample the node type w.r.t the distribution and represent it as a one-hot vector $T_v$.
We update the discrete representation of a node $h_v$ according to its type based on the operation $\mathcal{U}$ in \eqref{eq.graph_embedding}.
Besides the discrete representation $h_v$, we add a node with the sampled type to the neural network architecture $\xx$.

\textbf{2) Add incoming edges.} Similarly, to compute the probability of directed edges $\{u \to v\}$ ($u$ is a node before the node $v$), we employ an another MLP base on the representation $h_u$ and $h_v$. and sample the edges.
Using on the incoming edges, we then update the representation $h_v$ by updating the aggregation of the incoming edges $\mathcal{A}(\{h_u:u \to v\})$ in \eqref{eq.graph_embedding}.
Meanwhile, we build connections to the current node in the architecture $\xx$ according to the sampled edges.

At the very beginning, we generate the input node by directly setting $T_v$ as the input and use $h_0$ as the initial state of the input node. 
Meanwhile, there are no incoming edges thus we set the aggregation as an all-zeros vector.
The input node is added into the network architecture $\xx$ as well.

In fact, the decoder is the generate model $p_\theta(h_{out} | \ss)$ in the evidence lower bound \eqref{eq.lower_bound}.
The first term in \eqref{eq.lower_bound} $\mathbb{E}_{z \sim q_{\boldsymbol \phi}(\ss|\xx)} \left[\log p_{\boldsymbol \theta}(\xx | \ss)\right]$ estimates the reconstruction accuracy, where the maximization of this term urges the high similarity between the primal DAG and the decoded graph. 
We evaluate the reconstruction accuracy by summing up the similarity of nodes and edges between the primal DAG and the generated graph, where the reconstruction accuracy is denoted as:
\begin{align}
  \label{eq.reconstruction_acc}
  \mathcal{L}_{rec} = \mathcal{L}_n + \mathcal{L}_e.
\end{align}

\section{Training and Inference}
\begin{figure}
  \centering
  \includegraphics[width=\textwidth]{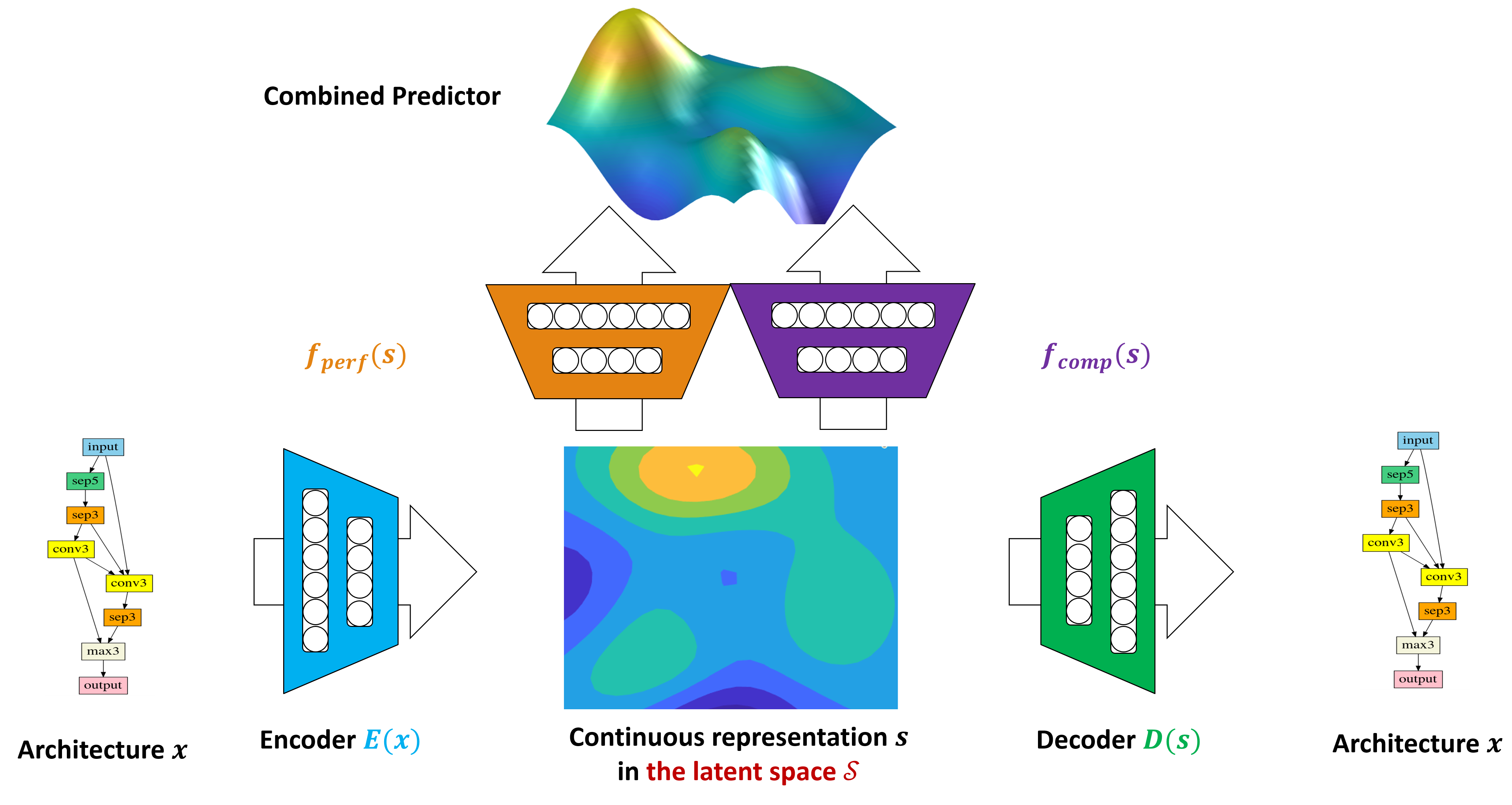}
  \caption{The training process of \texttt{NGAE}. \texttt{NGAE} jointly optimize the encoder $E: \mathcal{X} \to \mathcal{S}$, the decoder $D: \mathcal{S} \to \mathcal{X}$, the performance predictor $f_{perf}: \mathcal{S} \to \mathcal{Y}$ and the complexity predictor $f_{comp}: \mathcal{S} \to \mathcal{Z}$ via maximizing the objective \eqref{eq.training_objective}.}  
  \label{fig.training}
\end{figure}

To facilitate the training of the search models, we optimize the encoder, two predictors and the decoder in an end-to-end manner.
The objective of VAE is to maximize evidence lower bound in \eqref{eq.lower_bound}, while the objective of two predictors is to minimize the squared loss.
Therefore, to jointly learn those components, we combine the minimization of the KL divergence \eqref{eq.kl_divergence}, the squared loss \eqref{eq.squared_loss} and the reconstruction loss \eqref{eq.reconstruction_acc}, aiming to minimize the following objective:
\begin{equation}
  \label{eq.training_objective}
  \begin{aligned}
    \argmin\limits_{\phi, \theta, f_{perf}, f_{comp}} \quad  \mathcal{L} = \sum_{(\xx, y, z) \in (X, Y, Z)} 
    &~\mathcal{L}_n + \mathcal{L}_e + D_\text{KL} \big[q_\phi(\ss|h_{out}) ~||~ p_\theta(\ss)\big] \\
    + ~&\big[ f_{perf}(\ss) - y \big]^2 + \big[ f_{comp}(\ss) - z \big]^2,
  \end{aligned}
\end{equation}
where $\phi, \theta, f_{perf}, f_{comp}$ are the parameters for the encoder, the decoder, the performance predictor and the complexity predictor, respectively.
$\mathcal{L}_n$ and $\mathcal{L}_e$ measure the reconstruction loss in terms of nodes and edges.
Generated by the encoder, $h_{out}$ and $\ss$ are the discrete representation and the latent continuous representation for the architecture $\xx$, respectively.
Figure \ref{fig.training} illustrates the training process of \texttt{NGAE}, which outputs the latent space $\mathcal{S}$ and the objective regression model.

After the training process, we obtain a latent space $\mathcal{S}$ and two regression models $f_{perf}$ and $f_{comp}$ on this continuous space.
To find the architecture with optimal performance and high computational efficiency, we maximize the performance predictor and minimize the complexity predictor:
\begin{align}
  \label{eq.inference_objective}
  \argmax\limits_{\ss \in \mathcal{S}} \quad f(\ss) = \quad f_{perf}(\ss) - f_{comp}(\ss).
\end{align}

To solve the optimization problem in \eqref{eq.inference_objective}, we sample a point $\ss \in \mathcal{S}$ according to the prior $p_\theta(\ss) = \mathcal{N}({\boldsymbol 0}, \mathbf{I})$.
We use the first-order stochastic gradient method to optimize \eqref{eq.inference_objective}, and the continuous representation $\ss$ is updated along the direction of the gradient:
\begin{align*}
  \ss^{l+1}= \ss^{l} + \eta \frac{\partial f}{\partial \ss^{l}}, \qquad \text{where} ~~ l \in \{1, \cdots, L\},
\end{align*}
where $L$ is the maximal iterations.
After the inference, we use the decoder $D: \mathcal{S} \to \mathcal{X}$ to construct a valid neural architecture from the optimized representation $\ss'$ according to the generative mode $p_\theta(\xx|\ss)$.
So, the discovered architecture is 
$\xx' = D(\ss').$

\begin{algorithm}[t]
  \caption{Neural Architecture Optimization with Graph Embedding}
  \label{alg.NGAE}
  \begin{algorithmic}[1] % The number tells where the line numbering should start
      \Require The training data $(X, Y)$ with the architectures and their performance.
      The initialized encoder $E$, performance predictor $f_{perf}$, complexity predictor $f_{comp}$ and decoder $D$.
      Step size $\lambda$, batch size $b$ and number of iterations $T$ for training.
      Step size $\eta$ and number of iterations $L$ for testing.
		  \Ensure The neural architecture $\xx'$ with the optimal performance.
      \Procedure{Training}{} \Comment{Optimize $D, f_{perf}, f_{comp}, E$ by maximizing the objective \eqref{eq.training_objective}}
          \For{$t = 1, \cdots, T$}
              \State Split the training data $(X, Y)$ with batch size $b$.
                  \For{each batch $(X_i, Y_i)$} \Comment{Use sgd on each batch $(X_i, Y_i)$}
                      \State $\phi \gets \phi - \lambda \frac{\partial \mathcal{L}}{\partial \phi}$  \Comment{$\phi$ is the parameter set of the encoder $D$}
                      \State $f_{perf}  \gets f_{perf} - \lambda \frac{\partial \mathcal{L}}{\partial f_{perf}}$
                      \State $f_{comp}  \gets f_{comp} - \lambda \frac{\partial \mathcal{L}}{\partial f_{comp}}$
                      \State $\theta \gets \theta - \lambda \frac{\partial \mathcal{L}}{\partial \theta}$  \Comment{$\theta$ is the parameter set of the encoder $D$}
                  \EndFor
          \EndFor
          \State \textbf{return} the optimized $D, f, E$
      \EndProcedure

      \Procedure{Inference}{} \Comment{Optimize $\xx'$ by maximizing the objective \eqref{eq.inference_objective}}
          \State Sample a point $\ss' \in \mathcal{S}$ w.r.t $p_\theta(\ss)$.
          \For{$l = 1, \cdots, L$}
              \State $\ss'= \ss' + \eta \frac{\partial f}{\partial \ss'}$
          \EndFor
          \State Decode $\ss'$ to a architecture $\xx' = D(\ss')$
          \State \textbf{return} the discovered architecture $\xx'$
      \EndProcedure
  \end{algorithmic}
\end{algorithm}

We complete the detailed algorithm as shown in Algorithm \ref{alg.NGAE}.
In the training, the algorithm optimizes the encoder $E$, the performance predictor $f_{perf}$, the complexity predictor $f_{comp}$ and the decoder $D$ by maximizing the objective \eqref{eq.training_objective}, where we use the stochastic gradient descent (sgd) to solve the optimization.
In the inference, the algorithm starts with a sampled point $\ss'$ w.r.t the prior $p_\theta(\ss)$ and then optimize $\ss'$ by maximizing the performance and computational efficiency $f$ \eqref{eq.inference_objective}.

\section{Experiments}
Our experiments aim to estimate the effectiveness of the proposed framework in the neural architecture search domain, covering three complementary tasks:

{1) \bf Basic properties of the variational graph VAE.} The graph VAE is the core of our approach, which determines the continuous latent space and the ability of representations for architectures.
So, we first estimate the convergence of the reconstruction loss and KL divergence.
Finally, the smoothness of the latent space is illustrated by the visualization in the 2D subspace.

{2) \bf Predictive ability of the predictor $f$.} The predictor $f$ includes two nonlinear regression models that fit the performance and the complexity of architectures.
We optimize architectures in the latent space by maximizing the predictor $f$, thus the fitness of $f$ is significant to explore better architectures.
We compare the predicted performance and ground truth for randomly sampled graphs, and estimate the rooted-mean-squared error (RMSE) of $f$ in training and testing.

{3) \bf The comparison of the generalization ability and efficiency.}
Towards the generalization ability and efficiency, we compared the end-to-end framework \texttt{NGAE} with/without the complexity predictor and D-VAE.
DAG variational autoencoder (D-VAE) \cite{zhang2019d} first learns the VAE to obtain the latent space and then use Bayesian optimization to generate new architectures.

Different from the traditional NAS methods, our approach focus on powerful representations of neural architectures and the predictive model, thus the proposed methods don't need to train networks in the training. 
Here, our training dataset $(X, Y, Z)$ consists of 19, 020 neural architectures from the ENAS software \cite{pham2018efficient}.
The architectures include 6 nodes except the input node and the output node, where those nodes are sampled from 6 kinds of different operations: $3 \times 3$ convolution, $3 \times 3$ separable convolution, $5 \times 5$ convolution, $5 \times 5$ separable convolution, average pooling and max pooling. 
We split the dataset into $90\%$ training data and $10\%$ testing data, where we use the training data to learn the VAE and the predictor and testing data for evaluation.

\subsection{Basic properties of the variational graph VAE}

\begin{figure}
  \subfigure{\centering
  \includegraphics[width=0.5\textwidth]{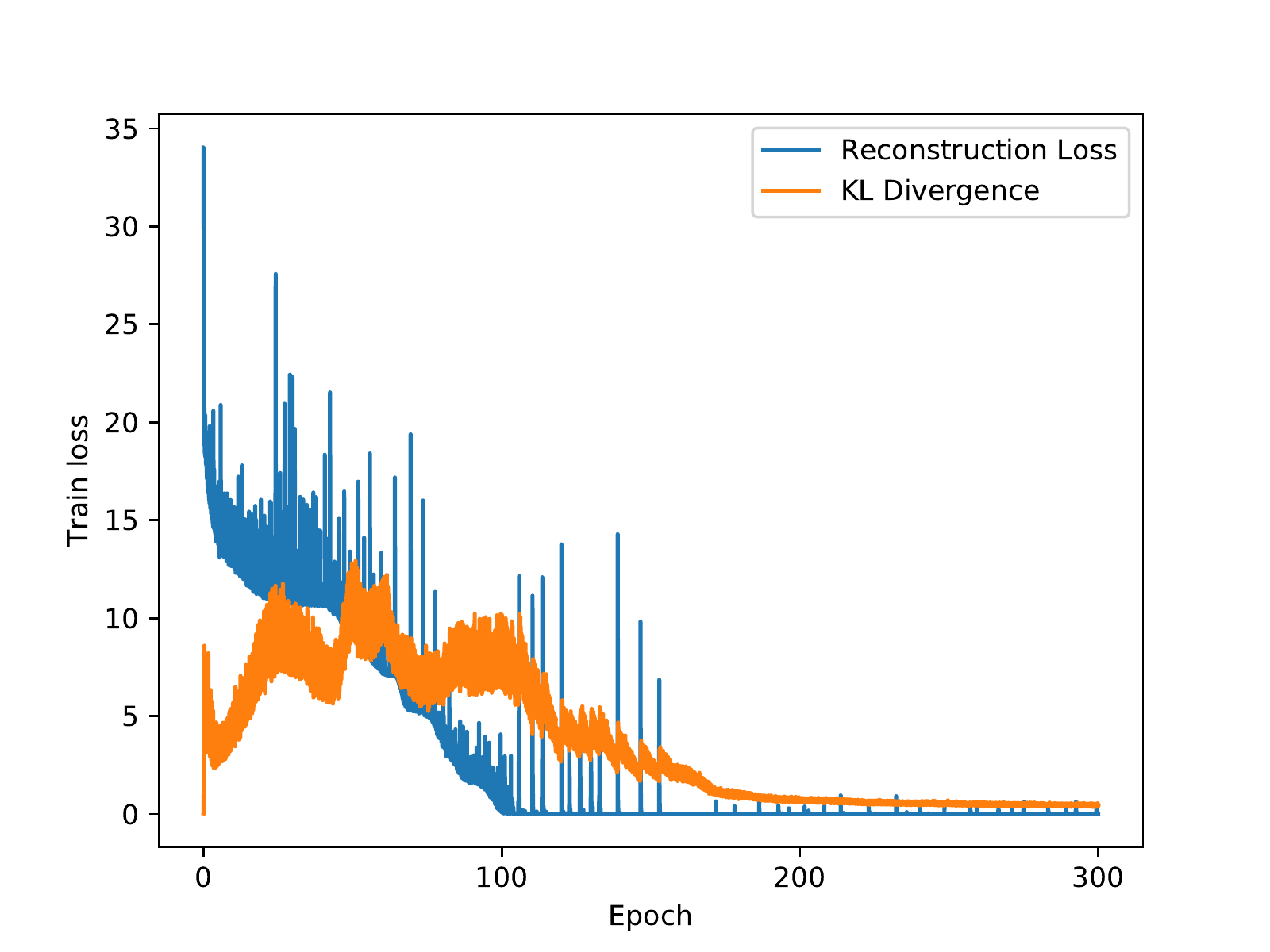}}
  \subfigure{\centering
  \includegraphics[width=0.5\textwidth]{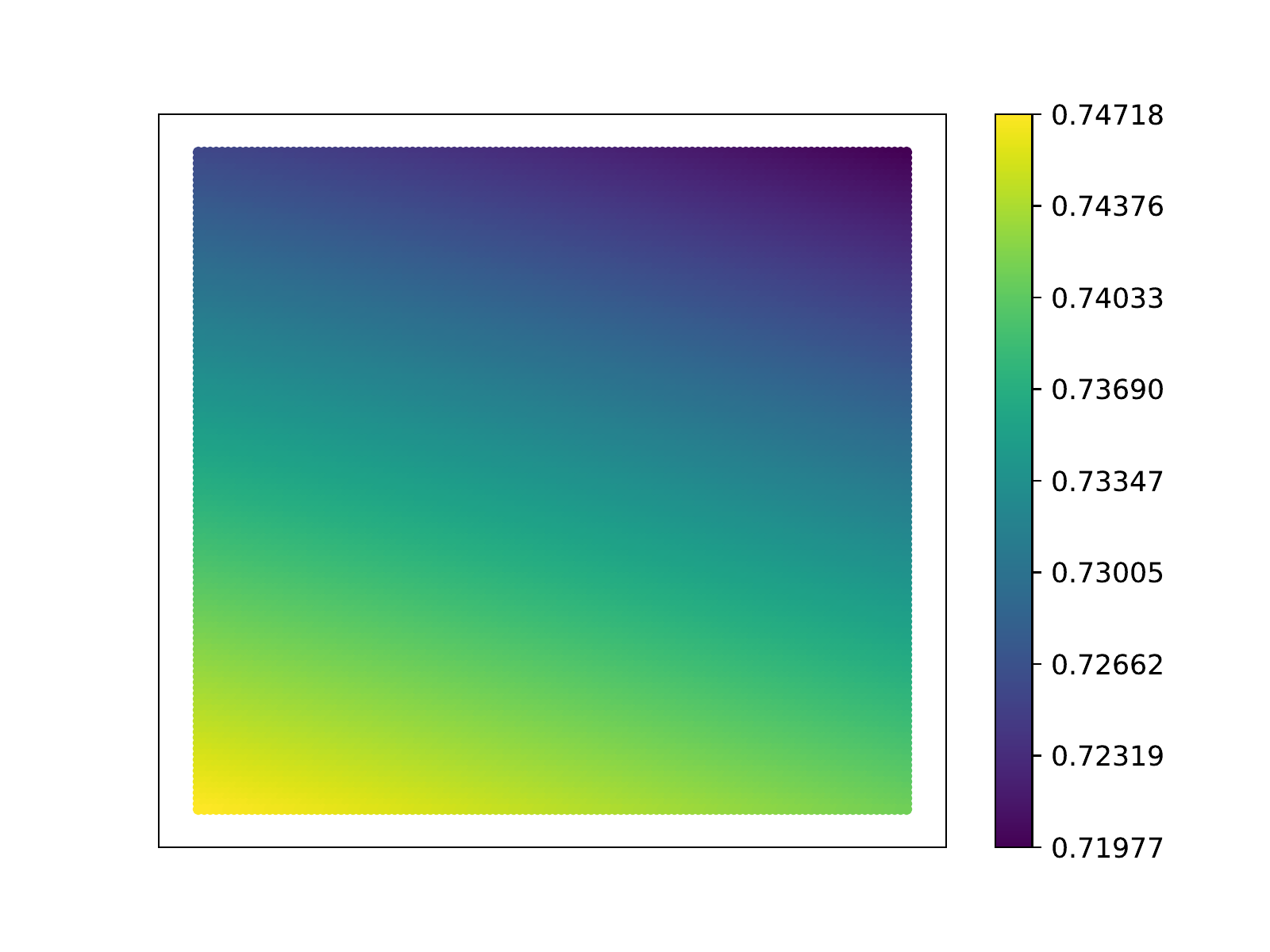}}
  \caption{The basic properties of the proposed approach. (left) The reconstruction loss and KL divergence during the training process. (right) Visualization of 2D subspace of the latent space.}
  \label{fig.exp1_properties}
\end{figure}

During the training process, we record the reconstruction loss $\mathcal{L}_n + \mathcal{L}_e$ and the KL divergence $D_\text{KL} \big[q_\phi(\ss|h_{out}) ~||~ p_\theta(\ss)\big]$.
The left figure in Figure \ref{fig.exp1_properties} reports the convergence of the reconstruction loss and the KL divergence that can be summarized as follows: 
(1) The reconstruction loss plays a dominant role in the training loss \eqref{eq.training_objective}, which reflects the match between the encoder $E$ and the decoder $D$. 
(2) The reconstruction loss is the main optimization term and decreases dramatically during the training. It convergence to a small constant near zero after 200 epochs.
(3) At the beginning, the encoder is relatively simple where the posterior approximation $q_\phi(\ss|h_{out})$ is close to the prior $p_\theta(\ss)$, thus the KL divergence is small.
As the encoder becomes more complex, the KL divergence grows while the reconstruction loss decreases. 
(4) After the reconstruction loss no longer reduces (100 epochs), the KL divergence begins to reduce to near zero.

\begin{table}[t]
  \centering
  \setlength\extrarowheight{1.3pt}
  \begin{tabular}{@{\extracolsep{0.2cm}}l|l|l|l}
     \hline
     Measures        & 100 epoch          & 200 epoch             & 300 epoch \\ \hline
     Accuracy        & 82.83          & 99.97             & 99.99 \\ \hline
     Validity        & 99.98          & 100             & 100 \\ \hline
     Uniqueness        & 51.55          & 42.05             & 38.56 \\ \hline
     Novelty        & 100           & 100              & 100  \\ \hline

  \end{tabular}
  \vspace{10pt}
  \caption{Reconstruction accuracy, prior validity, uniqueness and novelty ($\%$) for different periods.}
  \label{tab.performance_measures}
\end{table}

Using the optimized models under 100/200/300 epochs, we evaluate the performance of the proposed approach on testing dataset via four measures: 
(1) The \textit{reconstruction accuracy} estimates the rate of perfectly recovered graphs via graph VAE. 
(2) The \textit{validity} measures the proportion of valid DAGs generated from the prior distribution $p_\theta(\ss)$.
(3) The \textit{uniqueness} show the frequency of unique graphs from valid DAGs.
(4) The \textit{novelty} measures the proportion of novel graphs from valid DAGs which are not in the training set.
Table \ref{tab.performance_measures} reports the performance measures after 100/200/300 epochs. From Table \ref{tab.performance_measures}, we find (1) reconstruction accuracy is growing from $82.83\%$ and finally achieves $99.99\%$, (2) uniqueness declines as the training epochs increases because many optimized architectures tend to be same, (3) prior validity and novelty remain unchanged around $100\%$.

To estimate the smoothness of our method, we use the first two principal components of a continuous representation $\ss$ sampled by $p_\theta(\ss)$.
We then obtain more continuous representations by moving $\ss$ in the first two principal components and measure their performance predicted by the predictor $f$.
Results are reported in the right of Figure \ref{fig.exp1_properties} where two dimensionalities correspond to two principal components and the color bar indicates the predicted accuracies.
In the right of Figure \ref{fig.exp1_properties}, the predicted accuracy (the color) changes smoothly as the move of two principal components, showing that the generated latent space is continuous thus it can be optimized by gradient methods easily.

\subsection{Predictive ability of the predictor $f$}
\begin{figure}  
  \subfigure{\centering
  \includegraphics[width=0.33\textwidth]{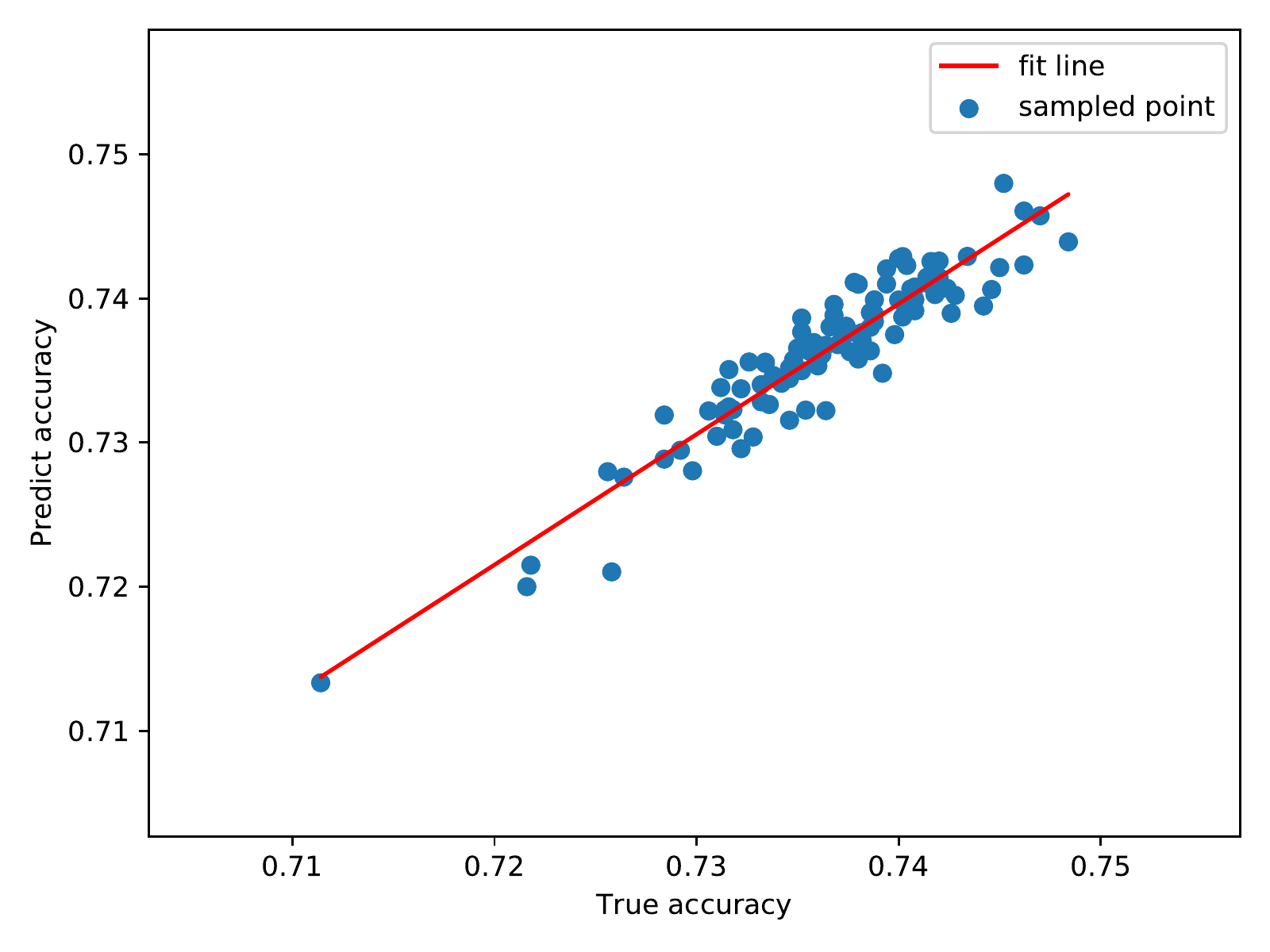}}
  \subfigure{\centering
  \includegraphics[width=0.33\textwidth]{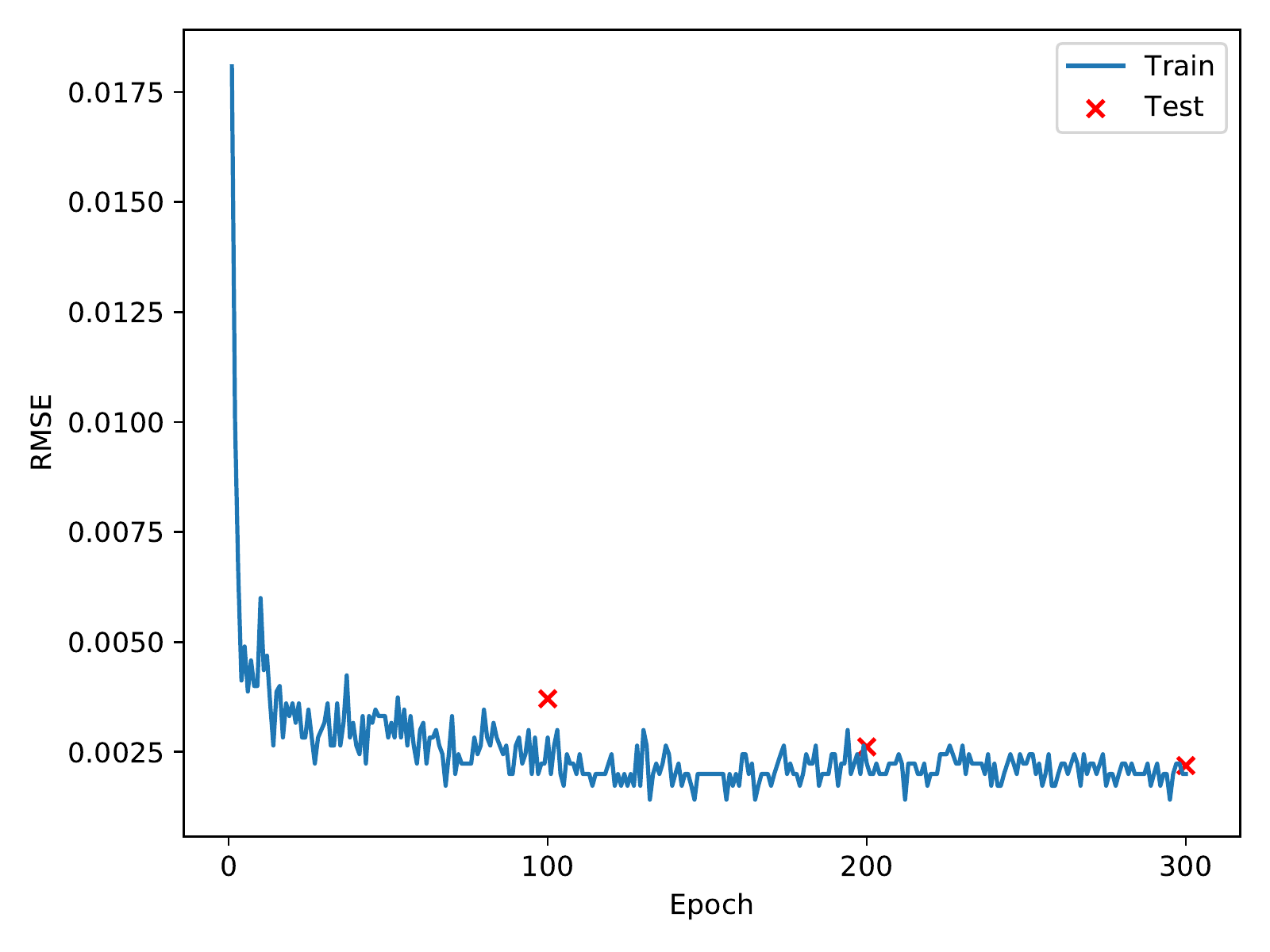}}
  \subfigure{\centering
  \includegraphics[width=0.33\textwidth]{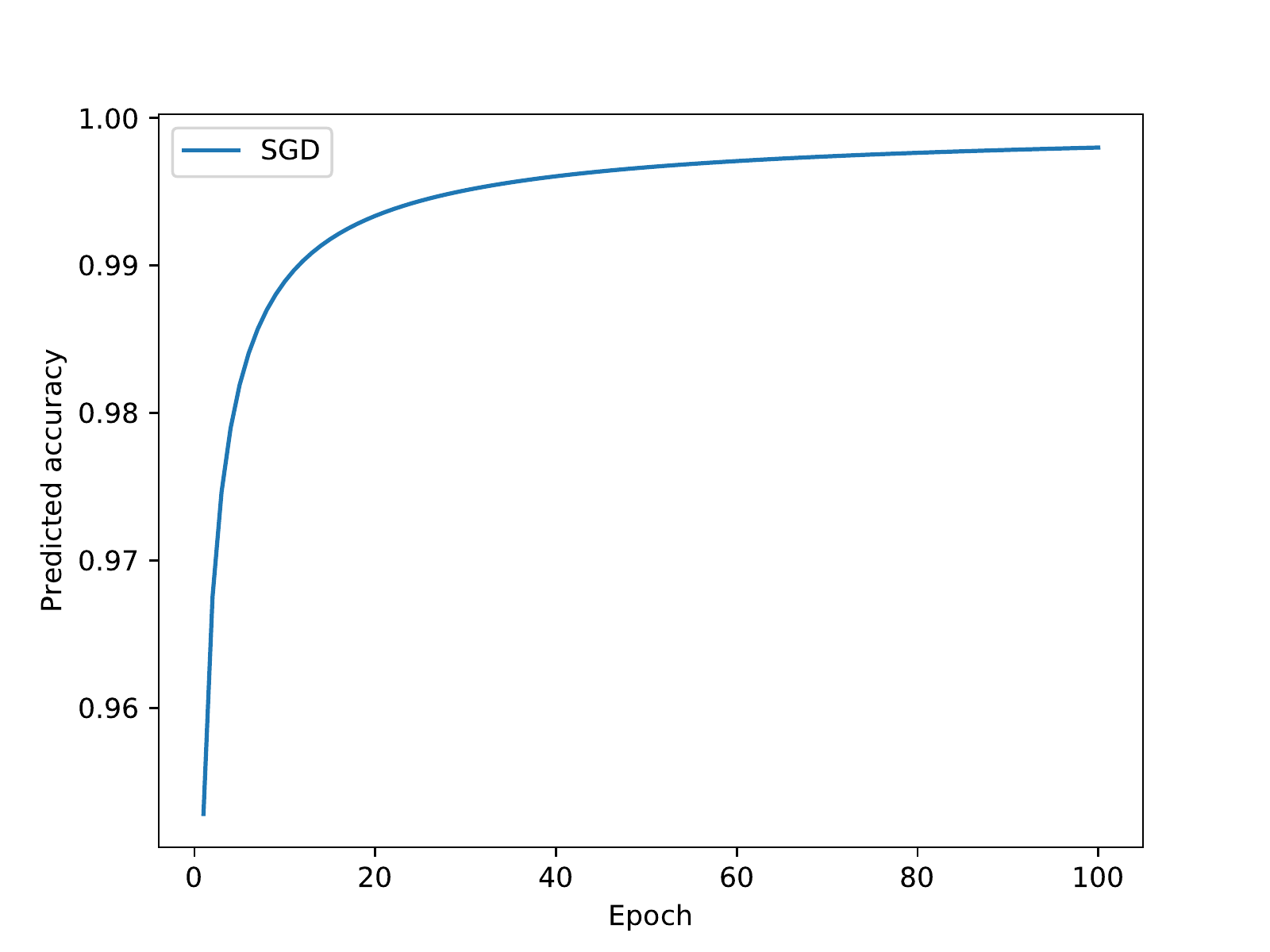}}
  \caption{Predictive performance of the predictor $f$. (left) True accuracies and predicted accuracies on sampled test architectures. (medium) The convergence of RMSE on training data and test data. (right) The predicted accuracy during inference using SGD.}
  \label{fig.exp2_predictor}
\end{figure}
In the left of Figure \ref{fig.exp2_predictor}, we record true accuracies and predicted accuracies of randomly sampled 100 architectures on the test set. 
The fit line is close to the line $y_{true} = y_{pred},$ that means our predictor $f$ achieve excellent predictive ability on accuracy.
The middle of Figure \ref{fig.exp2_predictor} illustrates the convergence of RMSE on training data and test data, where we only evaluate test RMSE on 100/200/300 epochs. 
RMSE convergent to a small number, improving the performance of $f$.
In the inference, we use the stochastic gradient ascent method to optimize the continuous representations by maximizing the output of predictor $f$.
The update of predicted accuracy in the inference is recorded in the right of Figure \ref{fig.exp2_predictor}.
The convergence of predicted accuracy reveals the superiority of multiple steps update of gradient over one-step update in NAO \cite{luo2018neural}.

\subsection{The comparison of the generalization ability and efficiency}
\begin{figure}[t]
  \centering
  \includegraphics[width=\textwidth]{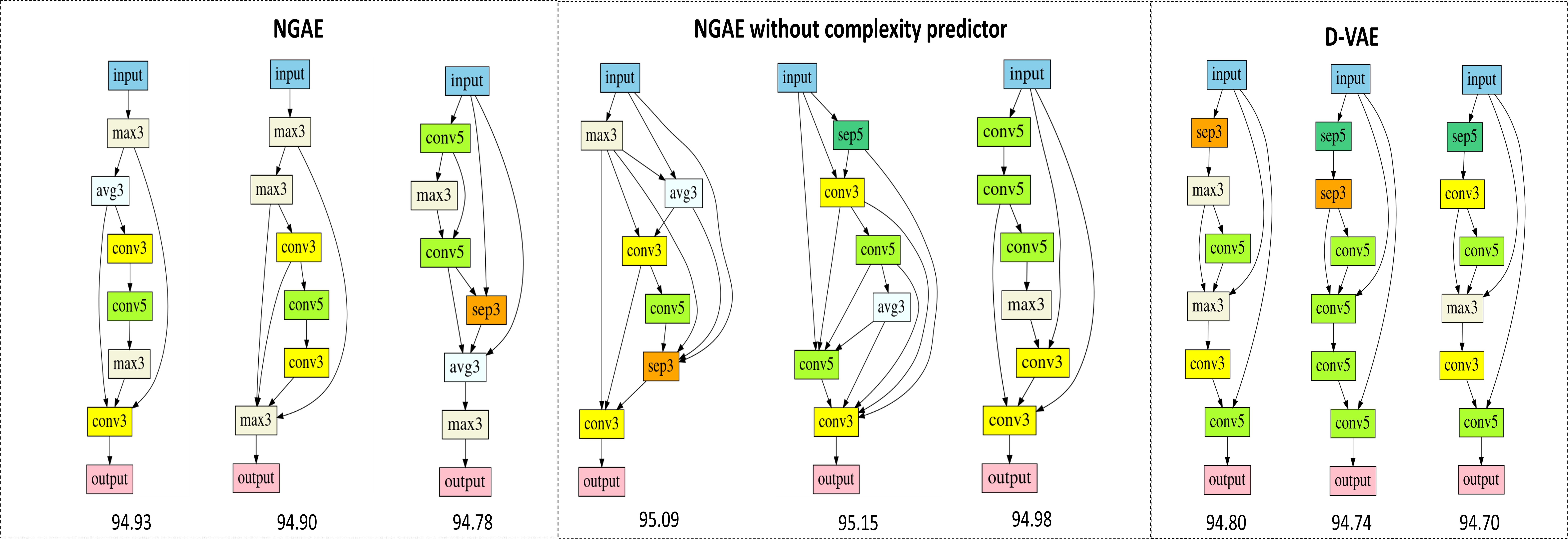}
  \caption{Top 3 architectures found by \texttt{NGAE}, \texttt{NGAE} without complexity predictor and D-VAE.}
  \label{fig.found_architectures}
\end{figure}

Figure \ref{fig.found_architectures} reports the top 3 neural architectures found by \text{NGAE}, \text{NGAE} without complexity predictor and D-VAE \cite{zhang2019d} on CIFAR-10.
\text{NGAE} without complexity predictor is a simpler version of \texttt{NGAE}, which only use the encoder, the performance predictor, the decoder in the framework, waiving the complexity predictor.
D-VAE use similar encoder and decoder to \texttt{NGAE}, but D-VAE only learns the encoder and decoder in representations training.
Then D-VAE trains a Bayesian optimization model and use it to search architectures.
\texttt{NGAE} and D-VAE apply similar graph VAE to construct a connections space, but \texttt{NGAE} learns the VAE and the optimizer (two predictors) in an end-to-end manner, while D-VAE split the training of VAE and the optimizer (Bayesian model).

From Figure \ref{fig.found_architectures}, we find that (1) Both \text{NGAE} with/without complexity predictor outperform D-VAE, proving the effectiveness of jointly training VAE and predictors. (2) \text{NGAE} without complexity predictor found architectures with best results but high computational complexity. (2) \text{NGAE} reduce computational complexity significantly by sacrificing little accuracy.

% The best architecture found by \texttt{NGAE} is $95.15\%$ on CIFAR-10, which falls behind the SOTA results (e.g. $97.89\%$ in NAONET + CUTOUT \cite{luo2018neural}).
% The reasons are as follows: (1) The outputs of \texttt{NGAE} are determined by training data. We use 6-layers networks with their accuracies from the ENAS software \cite{pham2013fast} as training set, where the search space is much smaller.
% If \texttt{NGAE} is performed on larger search space with deeper architectures, \texttt{NGAE} will find better architectures.
% (2) For the sake of fair comparison, our experiments haven't performed popular techniques such like cutout and more filters.
% (3) We combines the training process of VAE and predictors, boosting the efficiency of training, where we only use 1 GPU to train them in few hours.

\section{Conclusion and Future Work}
In this paper, we have proposed \texttt{NGAE}, using GNN-based VAE to construct a continuous space and gradient-based optimization to search new architectures on it.
There are two predictors (performance and complexity) built upon the latent space using the training set.
The combination of those predictors is used to find architectures with both better performance and lower computational complexity.

Due to our limited computational resources, we can't perform complex NAS methods, such as NAO \cite{luo2018neural}, \cite{liu2018darts} and NASP \cite{yao2020efficient}.
Meanwhile, we can also train the proposed \texttt{NGAE} using recent NAS-Bench-101 \cite{ying2019bench} and NAS-Bench-201 \cite{yao2020efficient}.
\texttt{NGAE} can not only search convolutional network architectures but also can search other types of architectures (e.g. RNN, GNN), only need to use their corresponding performance measures instead of accuracies on CIFAR-10.
We leave more compared methods, larger training datasets and other types of architecture for future work.

% \section*{Broader Impact}
% The continuous {\bf representations} of neural architectures and the {\bf optimization} on them are two key elements in recent NAS research.
% For representation issues, we employ both GNN and VAE to obtain powerful continuous representations of architectures. 
% For optimization issues, we jointly learn the graph VAE and two predictors using gradient methods.
% Due to the use of performance predictor and complexity predictor, our approach discovers architectures with both excellent performance and low computational burdens.
% The training and inference of \texttt{NGAE} are effective and elegant, which may inspire more research on the gradient-based NAS domain.

\bibliographystyle{unsrt}
\bibliography{all}

\end{document}